\documentclass[conference]{IEEEtran}
\IEEEoverridecommandlockouts
\usepackage{cite}
\usepackage{amsmath,amssymb,amsfonts}
\usepackage{graphicx, caption, subcaption}
\usepackage{textcomp}
\usepackage[table]{xcolor}
\usepackage{algorithm,algpseudocode}
\usepackage{hyperref}
\hypersetup{
     colorlinks = true,
     citecolor = green,
     }

\usepackage{multirow}
\usepackage{booktabs}
\usepackage[usestackEOL]{stackengine}

\def\BibTeX{{\rm B\kern-.05em{\sc i\kern-.025em b}\kern-.08em
    T\kern-.1667em\lower.7ex\hbox{E}\kern-.125emX}}
\begin{document}


\newboolean{showcomments}
\setboolean{showcomments}{true} 
\ifthenelse{\boolean{showcomments}}
  {
		\newcommand{\nbb}[2]{
		\fcolorbox{black}{yellow}{\bfseries\sffamily\scriptsize#1}
		{\sf$\blacktriangleright$\textcolor{blue}{\textit{#2}}$\blacktriangleleft$}
		}
		\newcommand{\version}{\emph{\scriptsize$-$9.2.2011$-$}}
		\newcommand{\remarks}[1]{\color{red}[#1]\color{black}}
		\newcommand{\copied}[1]{\color{green}[#1]\color{black}}
		\newcommand{\modified}[1]{\color{blue}[#1]\color{black}}
		\newcommand{\raw}{$\rightarrow$}
		\newcommand{\ins}[1]{\textcolor{blue}{\uline{#1}}} 
		\newcommand{\del}[1]{\textcolor{red}{\sout{#1}}} 
		\newcommand{\chg}[2]{\textcolor{red}{\sout{#1}}{\raw}\textcolor{blue}{\uline{#2}}} 
		\newcommand{\ugh}[1]{\textcolor{red}{\uwave{#1}}} 
  }
  {
		\newcommand{\nbb}[2]{}
		\newcommand{\remarks}[1]{}
		\newcommand{\modified}[1]{#1}
		\newcommand{\copied}[1]{#1}
		\newcommand{\version}{}
		\newcommand{\ugh}[1]{#1} 
		\newcommand{\ins}[1]{#1} 
		\newcommand{\del}[1]{} 
		\newcommand{\chg}[2]{#2} 
  }

\newcommand{\jens}[1]{\nbb{Jens}{#1}}
\newcommand{\cbe}[1]{\nbb{CBe}{#1}}
\newcommand{\lars}[1]{\nbb{Lars}{#1}}
\newcommand{\sankar}[1]{\nbb{Sankar}{#1}}
\newcommand{\ce}[1]{\nbb{CE}{#1}}
\newcommand{\mb}[1]{\nbb{Markus}{#1}}
\newcommand{\comment}[1]{\nbb{Comment}{#1}}

\title{Performance Analysis of Out-of-Distribution Detection on Various Trained Neural Networks}

\author{
\IEEEauthorblockN{Jens Henriksson\IEEEauthorrefmark{1}, Christian Berger\IEEEauthorrefmark{2}, Markus Borg\IEEEauthorrefmark{3},\\ Lars Tornberg\IEEEauthorrefmark{4}, Sankar Raman Sathyamoorthy\IEEEauthorrefmark{5}, Cristofer Englund\IEEEauthorrefmark{3}}\\
\IEEEauthorblockA{\IEEEauthorrefmark{1}Semcon AB, Gothenburg, Sweden, Email: jens.henriksson@semcon.com}
\IEEEauthorblockA{\IEEEauthorrefmark{2}University of Gothenburg and Chalmers Institute of Technology, Sweden, Email: christian.berger@gu.se}
\IEEEauthorblockA{\IEEEauthorrefmark{3}RISE Research Institutes of Sweden AB, Lund and Gothenburg, Sweden, Email: \{markus, cristofer\}@ri.se}
\IEEEauthorblockA{\IEEEauthorrefmark{4}Machine Learning and AI Center of Excellence, Volvo Cars, Gothenburg, Sweden, Email: lars.tornberg@volvocars.com}
\IEEEauthorblockA{\IEEEauthorrefmark{5}QRTech AB, Gothenburg, Sweden, Email: sankar.sathyamoorthy@qrtech.se}}

\maketitle

\begin{abstract}
Several areas have been improved with Deep Learning during the past years. For non-safety related products adoption of AI and ML is not an issue, whereas in safety critical applications, robustness of such approaches is still an issue. A common challenge for Deep Neural Networks (DNN) occur when exposed to out-of-distribution samples that are previously unseen, where DNNs can yield high confidence predictions despite no prior knowledge of the input.

In this paper we analyse two supervisors on two well-known DNNs with varied setups of training and find that the outlier detection performance improves with the quality of the training procedure. We analyse the performance of the supervisor after each epoch during the training cycle, to investigate supervisor performance as the accuracy converges. Understanding the relationship between training results and supervisor performance is valuable to improve robustness of the model and indicates where more work has to be done to create generalized models for safety critical applications. 

\end{abstract}


\section{Introduction}
Deep Neural Networks (DNN) constitute a major challenge to safety engineers, especially for complex and safety-critical functionality such as Autonomous Driving. Autonomous Driving requires complex perception systems embodying DNNs for Computer Vision to process the enormous amounts of video, lidar, and radar data streams. However, while DNNs have revolutionized the field of computer vision~\cite{lecun_deep_2015}, their characteristics and development process are completely different compared to the conventional software addressed by the automotive safety standard ISO~26262 -- Functional Safety~\cite{international_organization_for_standardization_iso_2018}. The behaviour of a DNN is not explicitly expressed by an engineer in source code following the principle where a developer defines the algorithm based on a specification; instead the developer defines an architecture that learns the algorithm, where engineers use enormous amounts of data complemented with domain-specific labels to let the machine figure out an algorithm that exhibits the desired behavior for a given stimulus. For autonomous driving, this step usually comprises the collection, preprocessing, training, and evaluation of huge amounts of annotated camera, lidar, and radar data using Machine Learning (ML)~\cite{salay_analysis_2017,henriksson_automotive_2018}.

In January 2019, ISO/PAS~21448 -- Safety of the Intended Functionality (SOTIF) was published to meet the pressing industrial need of an automotive safety standard that is appropriate for ML~\cite{international_organization_for_standardization_iso_2019}. A PAS (Publicly Available Specification) is not an established standard, but a document that closely resembles what is planned to become a future standard, i.e., a PAS is published to speed up a standardization process in response to market needs. A PAS must be transformed to a standard within six years, otherwise it will be withdrawn\footnote{https://www.iso.org/deliverables-all.html}.

ISO~26262 and SOTIF are meant to be complementary standards: ISO~26262 covers ``absence of unreasonable risk due to hazards caused by malfunctioning behaviour'' ¨\cite{international_organization_for_standardization_iso_2018} by mandating rigorous development, which is structured according to the established V-model way-of-working. SOTIF, on the other hand, addresses ``hazards resulting from functional insufficiencies of the intended functionality''~\cite{international_organization_for_standardization_iso_2019}, e.g., miss classifications by an object detector in an automotive perception system, which is different from conventional malfunctions targeted by defect-oriented ISO~26262. 

SOTIF is not organized according to the V-model but around 1) known safe states, 2) known unsafe states, and 3) unknown unsafe states. Furthermore, SOTIF presents a process to minimize the two unsafe states, represented by a flowchart orbiting around requirements specifications for the functionality under development, where moving hazards from 3) $\rightarrow$ 2) and from 2) $\rightarrow$ 1) originates from hazard identification and hazard mitigation, respectively.  

Following the SOTIF process, we identify that autonomous driving in environments that differ from what a DNN of the automotive perception was trained for, constitutes a major hazard; in ML literature, this is referred to as dealing with out-of-distribution samples. In previous work, we have proposed various DNN supervisors~\cite{henriksson_automotive_2019} to complement autonomous driving by detecting such samples as part of a safety cage architecture~\cite{borg_safely_2019}, i.e., adding a reject option to DNN classifiers when input does not resemble the training data.

As part of a mitigation strategy for out-of-distribution camera data for example, we argue that the SOTIF process should result in a safety requirement mandating the inclusion of a DNN supervisor. However, only a few previous studies have presented structured evaluations of different DNN supervisors for image data; most previous work instead focuses on a single supervisor, e.g., \cite{bendale2016towards,liang2017enhancing} and \cite{hendrycks2018deep}.

In this paper we evaluate two supervisor algorithms: a baseline algorithm~\cite{hendrycks2016baseline} and OpenMax~\cite{bendale2016towards}. In this context, the supervisor provides an anomaly score of each input sent to the model to comply with the test setup defined in \cite{henriksson_automotive_2019}. Our evaluation method consists of applying the two supervisors on two established DNNs namely VGG16~\cite{DBLP:journals/corr/SimonyanZ14a} and DenseNet~\cite{DBLP:journals/corr/HuangLW16a}, which are both trained with different hyperparameters on the CIFAR-10 dataset~\cite{CIFAR10}. For the outlier set, the validation set from Tiny ImageNet (a subset of the ImageNet large scale dataset~\cite{ILSVRC15}) is used. More specifically, this paper addresses the following two research questions: 

\begin{itemize}
    \item How does the outlier detection performance change with improved training of the DNNs under supervision?
    \item How does overfitting of the DNNs under supervision affect the supervisor performance?
\end{itemize}


\hfill 

We find that the performance of the supervisor is almost linear with regards to the network performance. However, the design and tuning of supervisors can be a volatile development process, especially when applied at different stages during training where network parameters have been shifted. 

The rest of the paper is organized as follows: Section~\ref{sec:rw} introduces related work on automotive perception and DNN supervisors. In Section~\ref{sec:eval}, we describe our method to evaluate and compare DNN supervisors based on our recent work~\cite{henriksson_automotive_2019}. Section~\ref{sec:results} presents our results and in Section~\ref{sec:disc} we discuss our findings in the light of safety for autonomous driving. Finally, Section~\ref{sec:conc} concludes our paper and outlines directions for future work.



\section{Related Work}
\label{sec:rw}

Recent work on AI/ML has gained a lot of attention for break-through on large-scale pattern recognition for various application scenarios ranging from popular old board games, over text and image analysis, up to complete end-to-end solutions for autonomous driving. As these reported results appear fascinating on the one hand and underline the possible potential of AI/ML, commercializing such solutions into safe products is the apparent next challenge to overcome for a successful roll-out. Hence, recent works are dealing with how to test deep learning models for example. Fei et al.~\cite{deepXplore} and the succeeding study from Guo et al.~\cite{DLFuzz} suggest to evaluate testing coverage of models by evaluating neuron coverage. While such approaches show similarities to code, statement, or branch coverage in software testing as defined in ISO~26262~\cite{international_organization_for_standardization_iso_2018}, they fall short for evaluating the robustness of NNs as increased neuron coverage as achieved by artificially constructed test samples does not necessarily correlate with out-of-distribution samples that a NN might experience in reality. Furthermore, the improvements of these approaches reported by the authors are in the range of 2-3\%. A different approach is suggested by Kim et al.~\cite{GuidedDLTesting} who analyzed the influence of a particular stimulus to a NN with the purpose of providing support when designing specific test data to evaluate the performance of a NN.

Czarnecki and Salay postulate in their position paper \cite{Czarnecki_2018} a framework to manage uncertainty originating from perception-related components with the goal of providing a performance metric. Their work provides initial thoughts and concepts to describe uncertainty as present in ML-based approaches; however, in contrast to our work here, uncertainty originating from the trained model itself is only captured briefly and not in a quantitative approach as outlined here.

\begin{table}[!t]
    \centering
    \caption{Metrics for evaluating supervisor performance \cite{henriksson_automotive_2019}. }
    \label{tab:evaluation_metrics}
\begin{tabular}{| c | p{0.75\linewidth}|}
\hline \textbf{Metric} & \textbf{Description} \\
\hline AUROC  & Area under the Receiver Operating Characteristic (ROC) curve. Captures how the true/false positive ratios change with varying thresholds. \\
\hline AUPRC & Area under the Precision-Recall curve. Indicates the precision variation over increasing true positive rate.\\
\hline TPR05 & True positive rate at 5\% false positive rate. Determines the rise of the ROC-curve, with minimal FPR.  \\
\hline P95 & Precision at 95\% recall. Presents the accuracy when removing the majority of outliers. \\
\hline FNR95 & False negative rate at 95\% false positive rate. Shows how many anomalies are missed by the supervisor. \\
\hline CBPL & Coverage breakpoint at performance level. Measures how restrictive the supervisor has to be to regain similar accuracy as during training. \\
\hline CBFAD & Coverage breakpoint at full anomaly detection. Gives the coverage level where the supervisor has caught all outliers. \\
\hline
\end{tabular}
\end{table}


Regarding out-of-distribution detection, several recent works have been proposed. Hendrycks and Gimpel~\cite{hendrycks2016baseline} describe an approach to monitor the performance of a NN-based system by evaluating the Softmax layer. This is used to predict if the network is classifying a given sample correctly; in addition, they also use this feature as a discriminator between in-distribution and out-of-distribution from the training data. 

Similarly to Hendrycks and Gimpel, Liang et al.~\cite{Liang18} also use the Softmax layer in their approach named ODIN but without the need for any changes in a given model. Bendale et al.~\cite{bendale2016towards} propose a new model layer OpenMax, which is a replacement for the Softmax layer. This new layer compares the input sample towards an unknown probability element, adapted from meta-recognition of the penultimate layer, thus allowing the layer to learn the behavior of the training set, and correlated classes.  

Additionally, detection of adversarial perturbations has been a topic of growing interest to highlight vulnerabilities in neural networks. Metzen et al. \cite{metzen2017detecting} proposed to extend the neural net with sub-networks aimed specifically at detect adversarial perturbations. In contrast to related work that aims to increase robustness of the classification task, their work empirically shows that their sub-networks can be trained on specific adversarials and generalized to similar or weaker perturbations. 

Carlini and Wagner~\cite{carlini2017towards} show how adversarial attacks can break defensive distillation \cite{papernot2016distillation}, a recent approach tailored to reduce the success rate of adversarial attacks from 95\%  to 0.5\%. Their adversarial attack is tailored to three distance metrics, which are commonly used when generating adversarial samples. 

Finally, in contrast to only monitoring the Softmax layer, Tranheden and Landgren~\cite{MattiasLudwig18} compared and extracted information from five different detection methods to craft their own. Their combined solution suggests to supervise activations on all layers as various models and the problems being addressed vary quite a lot and hence, model-specific supervisors are needed.

\section{Evaluation Method} \label{sec:eval}
The scope of this work consists of training two image classification networks to various degree of accuracy. This is followed by applying two supervisors and documenting the change of supervisor performance. In addition, it is of interest to study how longer sessions of training the network affect the performance of the supervisors. To catch the behavioral change, the supervisor is applied after every $10^{th}$ epoch. The training process, the supervisor algorithms, and how the evaluation is carried out is explained in detail in the following. 

\textbf{Model training:} The experiments are completed with VGG16~\cite{DBLP:journals/corr/SimonyanZ14a} and DenseNet~\cite{DBLP:journals/corr/HuangLW16a}. VGG16 is a 16-layer deep network with all layers connected in series. DenseNet, a more modern network than VGG, consists of dense blocks of interconnected layers in between traditional convolutional layers. Both networks work in a feed-forward fashion, and are trained with the CIFAR-10 dataset~\cite{CIFAR10}. 

An interesting aspect of training DNNs is the continuous challenge to avoid overfitting. Over the years, several methods with the intention to improve generalization have been presented such as dropout layers~\cite{srivastava2014dropout}, batch normalization~\cite{ioffe2015batch}, methods of modifying learning rate or augmenting images with noise and shifts \cite{lecun2012efficient}. 

In this work, we study how the supervisor performance changes when training for longer periods of epochs, with three different levels of support to improve generalization. Thus, the networks are retrained three times; one in a normal fashion; one with the extension of image augmentation; one with the extension of image augmentation and adaptive learning rates and save states. 

\begin{algorithm}[!tpb] 
\caption{The \textit{Baseline} algorithm that utilizes the Softmax of the output vector to accept/reject~\cite{hendrycks2016baseline}}\label{Alg:Baseline}
\begin{algorithmic}[1]
\Require threshold $\epsilon$
\State Compute the output vector \textbf{v}(\textbf{x}) for input sample \textbf{x}
\State Let $P$ be the Softmax of output vector \textbf{v} for all classes $j = 1, .., N$
\Statex \begin{equation} P(y=j |\textbf{x})= \frac{ e^{\textbf{v}_{j}(\textbf{x})} }{\sum_{i=1}^{N} e^{\textbf{v}_{i}(\textbf{x})}} \end{equation}
\State Let the anomaly score be $A(\textbf{x})$ = 1 - argmax $P(y=j |\textbf{x})$ where $argmax$ is the function argument that maximizes function value.

\State Let discriminator $D\{0, 1\}$ be defined as \begin{equation}
    D = \left\{\begin{matrix}
\begin{matrix}
1 & if A(\textbf{x}) <  \epsilon \\ 
0 & otherwise
\end{matrix}
\end{matrix}\right.
\end{equation}

\end{algorithmic}
\end{algorithm}

\begin{algorithm}[!t] 
\caption{The \textit{OpenMax}~\cite{bendale2016towards} algorithm finds a Weibull distribution through Meta-Recognition calibration for each class $j$, with $\eta$ largest accepted distance to mean $\mu_{j}$. The rejection criterion first revises $\alpha$ top classes, while adding up distance as an ``unknown unknown'' probability. The sample is rejected if the most likely class is the unknown unknown or below threshold $\epsilon$.}\label{Alg:Openmax}
\begin{algorithmic}[tbh]
\Require Output vectors of the penultimate layer from the training set $\textbf{V}(x)=\textbf{v}_{1}(x) ... \textbf{v}_{N}(x) $
\Require Hyperparameter $\eta$: Largest accepted distance from sample to mean.
\Require Meta Recognition toolbox $LibMR$ for Weibull model fitting

    \Procedure{Fit Weibull model}{$\mathbf{V}(\mathbf{x}), \eta$}
      \State For each correct classification let $\textbf{S}_{i,j} =  \textbf{v}_{j}(x_{i,j})$ 
      \For{\texttt{each class j = 1 ... N}}
        \State Compute the mean output vector $\mu_{j} = mean(\textbf{S}_{i,j})$
        \State Fit Weibull model $\rho_{j} = libMR.fit(\hat{S}, \mu_{j}, \eta )$ 
      \EndFor
      \State \textbf{return} $\rho$, $\mu$ 
    \EndProcedure
    \Statex
    
    \Require Hyperparameter $\alpha$: How many possible classes to revise, sorted by their respective probability
    \State Compute output vector \textbf{v}(\textbf{x}) for input sample \textbf{x}
    \State Let $s(i) = argsort(v_{j}(x))$ and $\omega_{j}=1$
    \For{$i = 1, ..., \alpha$}
        \Comment{Recallibrate weibull score}
        \State $\omega_{s(i)} = \rho_{i} (\mathbf{v}(\mathbf{x}), \mu_{i})$
    \EndFor
    \State Revise output vector $\hat{v}(\mathbf{x}) = \mathbf{v}(\mathbf{x}) \cdot \omega (\mathbf{x})$ 
    \State Define  $\hat{v}_{o}(\mathbf{x}) = \sum_{i} \mathbf{v}_{i}(\mathbf{x}) (1-\omega_{i}(\mathbf{x}))$
    \Statex

    \Require threshold $\epsilon$
    
    \State Let $\hat{P}$ be the Softmax of output vector $\hat{\mathbf{v}}$ for all classes $j = 1, .., N$
    \Statex \begin{equation} \hat{P}(y=j |\textbf{x})= \frac{ e^{\hat{\textbf{v}}_{j}(\textbf{x})} }{\sum_{i=1}^{N} e^{\hat{\textbf{v}}_{i}(\textbf{x})}} \end{equation}
\State Let the anomaly score be $A(\textbf{x})$ = 1 - argmax $\hat{P}(y=j |\textbf{x})$ where $argmax$ is the function argument that maximizes function value.

\State Let the discriminator $D\{0, 1\}$ be defined as \begin{equation}
    D = \left\{\begin{matrix}
\begin{matrix}
1 & if A(\textbf{x}) <  \epsilon \text{ or argmax }\hat{P} == \hat{v}_{o}(\mathbf{x}) \\ 
0 & otherwise
\end{matrix}
\end{matrix}\right.
\end{equation}

\end{algorithmic}
\end{algorithm}

\textbf{Evaluation metrics:}
The metrics to evaluate the performance of the supervisors are selected based on our previous study~\cite{henriksson_automotive_2019}. The study derived and motivated seven metrics and four plots that describe the performance of a supervisor. The metrics present the supervisors' ability in discriminating samples from an out-of-distribution set, as well as the risk that the sample is wrongly classified depending on the desired level of coverage. All evaluation metrics are described in Tab.~\ref{tab:evaluation_metrics}. This paper uses the first six metrics, and excludes \textit{CBFAD}, since full anomaly detection is not expected from this study.\footnote{Since CIFAR-10 and Tiny ImageNet are not disjoint, it is unreasonable to expect full separability.}

\textbf{Supervisors:}
This study evaluates a state-of-the-art supervisor, together with a baseline algorithm. In the context of this paper, a supervisor is defined as a monitoring system that refers to an inlier/outlier as a negative/positive sample. The supervisor has access to the training data, the inputs, as well as network activations of the neural network to determine whether or not an input sample should be considered out-of-distribution compared to the accumulated knowledge in the model. A supervisor can be realized in different ways, such as solely comparing the input sample to the training data (variational autoencoders \cite{An2015VariationalAB}), utilizing network layer observations (Baseline \cite{hendrycks2016baseline}, OpenMax \cite{bendale2016towards}), or applying perturbations to the input sample (ODIN \cite{liang2017enhancing}, Generative Adversarial Networks \cite{goodfellow_nips_2016}). To comply with the evaluation metrics, the only requirement is that the supervisor provides a single value: the anomaly score, which measures the similarity of the sample to the accepted inlier domain.

This paper compares two methods: OpenMax~\cite{bendale2016towards}, and a baseline algorithm~\cite{hendrycks2016baseline}. The selected supervisors all consist of a manipulation, followed by a discriminate criterion, $D = \{0, 1\}$, that corresponds to if the supervisor accepts or rejects the sample. The methods are described in detail here: 

\textit{Baseline:}
The baseline supervisor utilizes the penultimate layer of a network, to extract the network activations for all classes, as described in Algorithm~\ref{Alg:Baseline}. The algorithm subtracts the Softmax value of the most probable prediction, thus giving high anomaly scores to samples with scattered predictions, i.e., there is no clear best prediction. 


\textit{OpenMax:} The \textit{OpenMax} supervisor works under the preconception that most categories have a relative consistent pattern for output vectors. The idea is that the combined information in the output vectors for a training class can be used to fit a probability distribution function. Each input sample can then be compared to the most likely distribution, and rejected if the distance is too far from it. The OpenMax supervisor catches fooling samples that are artificially constructed, since their output vectors are quite different compared to real images. 

Even though it is possible to add an extra class for \textit{known unknowns}, which means incorporating a larger variety of samples as an outlier class for a network, it is infeasible to train with regards to \textit{all} possible unknowns. Hence, tools that account for \textit{unknown unknowns}, or \textit{unknown unsafe states} as defined in the SOTIF specification \cite{international_organization_for_standardization_iso_2019}, are needed to diminish the cause of unknown inputs. \textit{OpenMax} does this by extending the categories by an \textit{unknown unknown} class, that incorporates the likelihood of the DNN to falsely classify a sample. To do this, OpenMax adapts the concept of Meta-Recognition to work for DNNs, see Algorithm~\ref{Alg:Openmax}.


\section{Experimental results} \label{sec:results}
Each network is trained in three different runs, where each run adds tools to increase generalization of the model. The three runs consist of a normal one with no added support; the second one uses an added preprocessing step where each image is manipulated with added noise or an image transformation such as a rotation, shift or crop; and the third one uses the same as the second one with the addition of variable learning rate, where the learning rate is changed after 100 and 200 epochs. To differentiate between these three runs, the results section include a naming extension for the latter two training runs, i.e.,  \textit{Augm.} and \textit{Augm. + lr}, respectively. 

\begin{figure}[!tpb] 
    \includegraphics[width=1\linewidth]{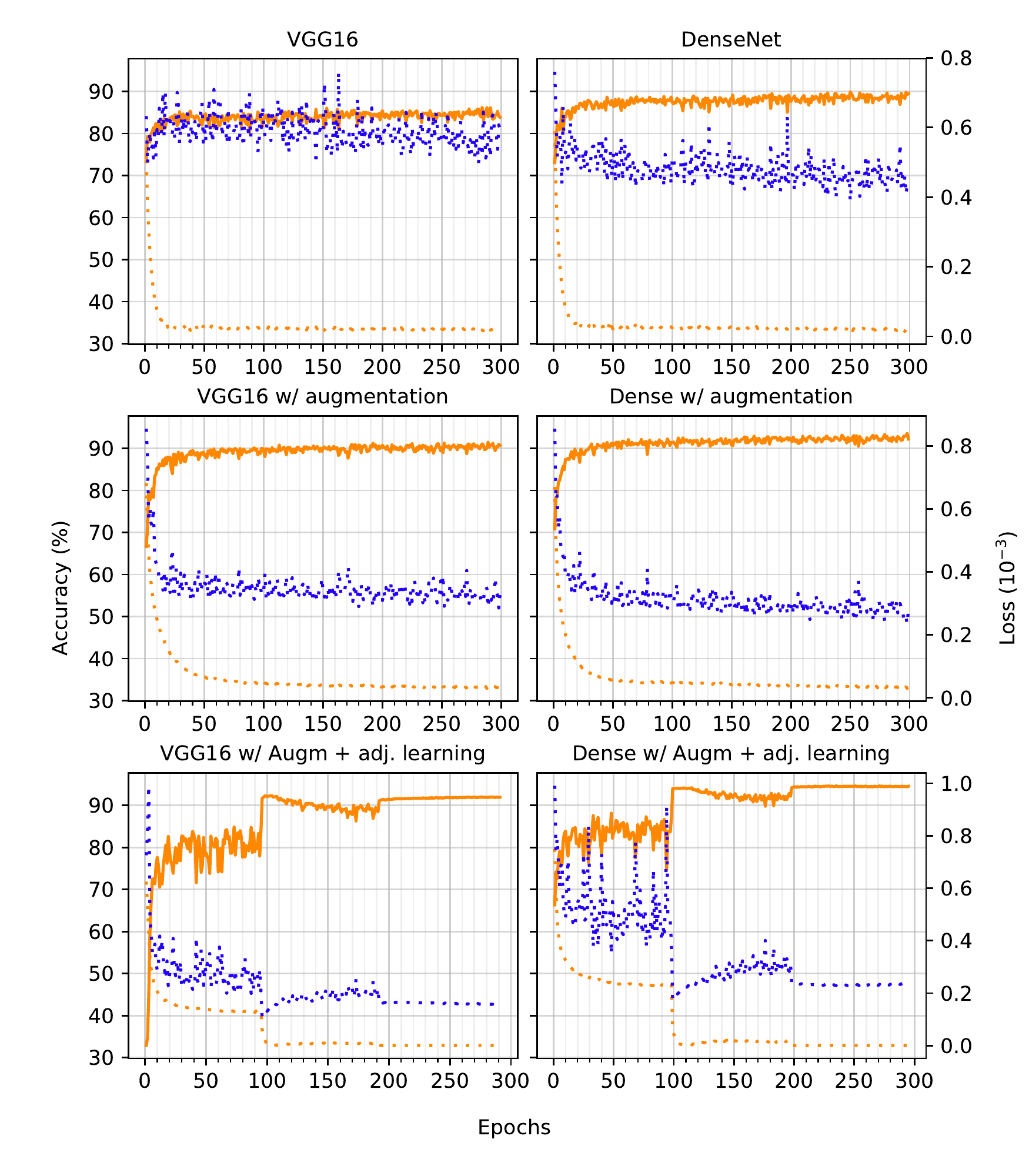}
    \caption{Training results from the six training runs with VGG16 to the left, and DenseNet to the right. The solid and dashed line represent accuracy and loss respectively. Colors orange and blue represent results on training and testing set respectively.}
    \label{fig:training}
\end{figure}

\begin{table*}[!t]
\centering
\caption{Results for VGG16 and DenseNet, over three separate training runs, after applying the two supervisors. Arrows indicate the desired direction of the value, i.e $\uparrow$ indicates higher value is better.} \label{tab:supervisor_best_results}
\begin{tabular}{lcllllllllll}
\hline
 & \multicolumn{5}{c}{\textbf{Training results}} & \multicolumn{6}{c}{\textbf{Metrics from \cite{henriksson_automotive_2019}}} \\ \hline
\textbf{Model} & \textbf{Augmented} & \textbf{Supervisor} & \textbf{Epoch} & \textbf{Acc} $\uparrow$ & \multicolumn{1}{l|}{\textbf{Loss} $\downarrow$} & \textbf{AUROC} $\uparrow$ & \textbf{AUPRC} $\uparrow$& \textbf{TPR05} $\uparrow$& \textbf{FNR95} $\downarrow$ & \textbf{P95} $\uparrow$& \textbf{CBPL} $\uparrow$ \\ \hline
\multirow{6}{*}{VGG16} 
 & - & Baseline & 291 & 86.11 & \multicolumn{1}{l|}{0.00622} & 0.80547 & 0.76597 & 0.20530 & 0.72246 & 0.00450 & 0.50430 \\
 & - & OpenMax & 291 & 86.11 & \multicolumn{1}{l|}{0.00622} &  0.80794 & 0.79987 & 0.00000 & 0.60538 & 0.00410 & \textbf{0.50940} \\
 & Augm. & Baseline & 293 & 91.20 & \multicolumn{1}{l|}{0.00334} &  0.84490 & 0.80894 & 0.27210 & 0.70305 & 0.00150 & 0.45435  \\
 & Augm. & OpenMax & 293 & 91.20 & \multicolumn{1}{l|}{0.00334} &  0.85337 & 0.83897 & 0.34800 & 0.61430 & 0.00120 & 0.46285  \\
 & Augm. + LR & Baseline & 291 & 91.95 & \multicolumn{1}{l|}{0.00367} &  0.81447 & 0.84310 & 0.00000 & 0.54946 & \textbf{0.01050} & 0.46010 \\
 & Augm. + LR & OpenMax & 291 & 91.95 & \multicolumn{1}{l|}{0.00367} &  0.83994 & 0.82207 & 0.32650 & 0.79910 & 0.00970 & 0.43205 \\ \hline
\multirow{6}{*}{DenseNet} 
 & - & Baseline & 297 & 89.84 & \multicolumn{1}{l|}{0.00409} &  0.82407 & 0.79411 & 0.25920 & 0.70025 & 0.00150 & 0.46610 \\  
 & - & OpenMax & 297 & 89.84 & \multicolumn{1}{l|}{0.00409} &  0.83312 & 0.82654 & 0.41430 & 0.60422 & 0.00230 & 0.47495 \\ 
 & Augm. & Baseline & 298 & 93.11 & \multicolumn{1}{l|}{0.00254} &  0.84762 & 0.81879 & 0.30630 & 0.70392 & 0.00240 & 0.41525 \\
 & Augm. & OpenMax & 298 & 93.11 & \multicolumn{1}{l|}{0.00254} &  0.85599 & 0.85198 & 0.38680 & 0.60769 & 0.00350 & 0.41635  \\
 & Augm. + LR & Baseline & 284 & 94.56 &  \multicolumn{1}{l|}{0.00231} & \textbf{0.87478} & 0.86075 & \textbf{0.41970} & 0.73911 & 0.00490 & 0.41845 \\
 & Augm. + LR & OpenMax & 284 & 94.56 &  \multicolumn{1}{l|}{0.00231} &  0.85966 & \textbf{0.87657} & 0.00000 & \textbf{0.54820} & 0.00550 & 0.42040 \\ \hline
\end{tabular}
\end{table*}

Each model is trained for 300 epochs, with a batch size of 64. This is more than enough to allow all models to converge. For the two initial runs the learning rate is fixed to $10^{-2}$, whereas for the final run the learning rate is varying between $10^{-1}$ and $10^{-2}$. The complete training results can be seen in Fig.~\ref{fig:training}. 

For both VGG16 and DenseNet, the addition of random augmentation of the input images allows the network to better generalize and to not get stuck in local minima. Augmentation is a method that is commonly used to do just this, as well as to reduce overfitting. 

For all our models, already after 10 epochs the test and training losses start to diverge from each other, which indicates that the model is learning faster towards the training set compared to the test set. While this is generally not desired, it is acceptable as long as the general performance on the test set also improves, which is the case for the remaining epochs. The only scenario where the test performance gets worse is when the model is allowed to adjust the learning rate, which intuitively means it will converge to a local optimum. This behavior can be seen in the bottom plots in Fig.~\ref{fig:training}, where the learning rate is reduced after 100 epochs to match the other networks. At this stage, it seems that the training starts overfitting. The learning rate is altered again after 200 epochs, but at this stage the learning rate is so small and hence, the network will only do small fine-tuning. 



\begin{figure}[!b] 
    \centering
    \includegraphics[width=1\linewidth]{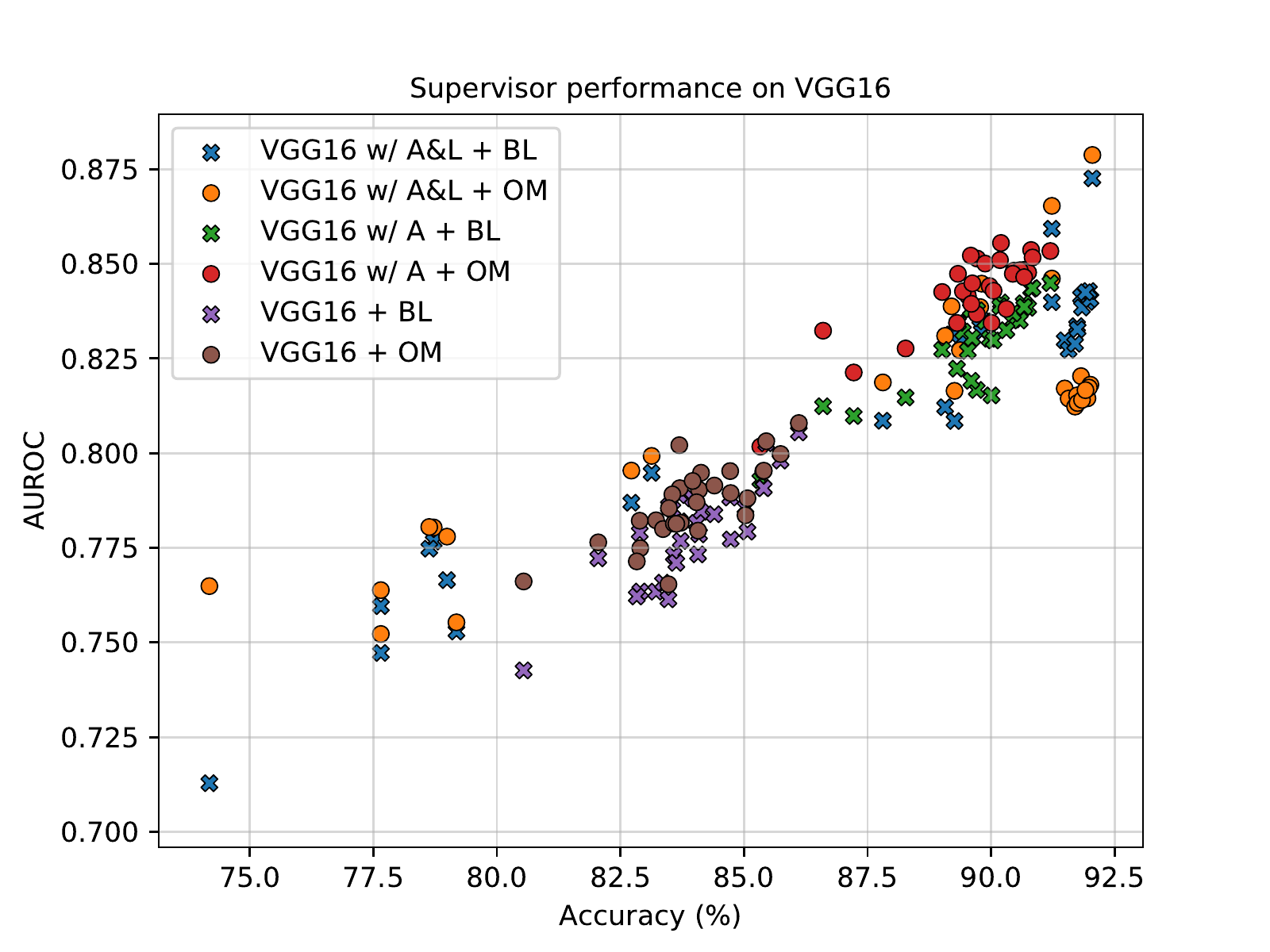}
    \caption{AUROC metrics compared to the accuracy of the VGG16 network. With higher network accuracy, the supervisor can easier distinguish between in- and outliers.}
    \label{fig:vgg-auroc-accuracy}
\end{figure}

\begin{figure}[!b] 
    \centering
    \includegraphics[width=1\linewidth]{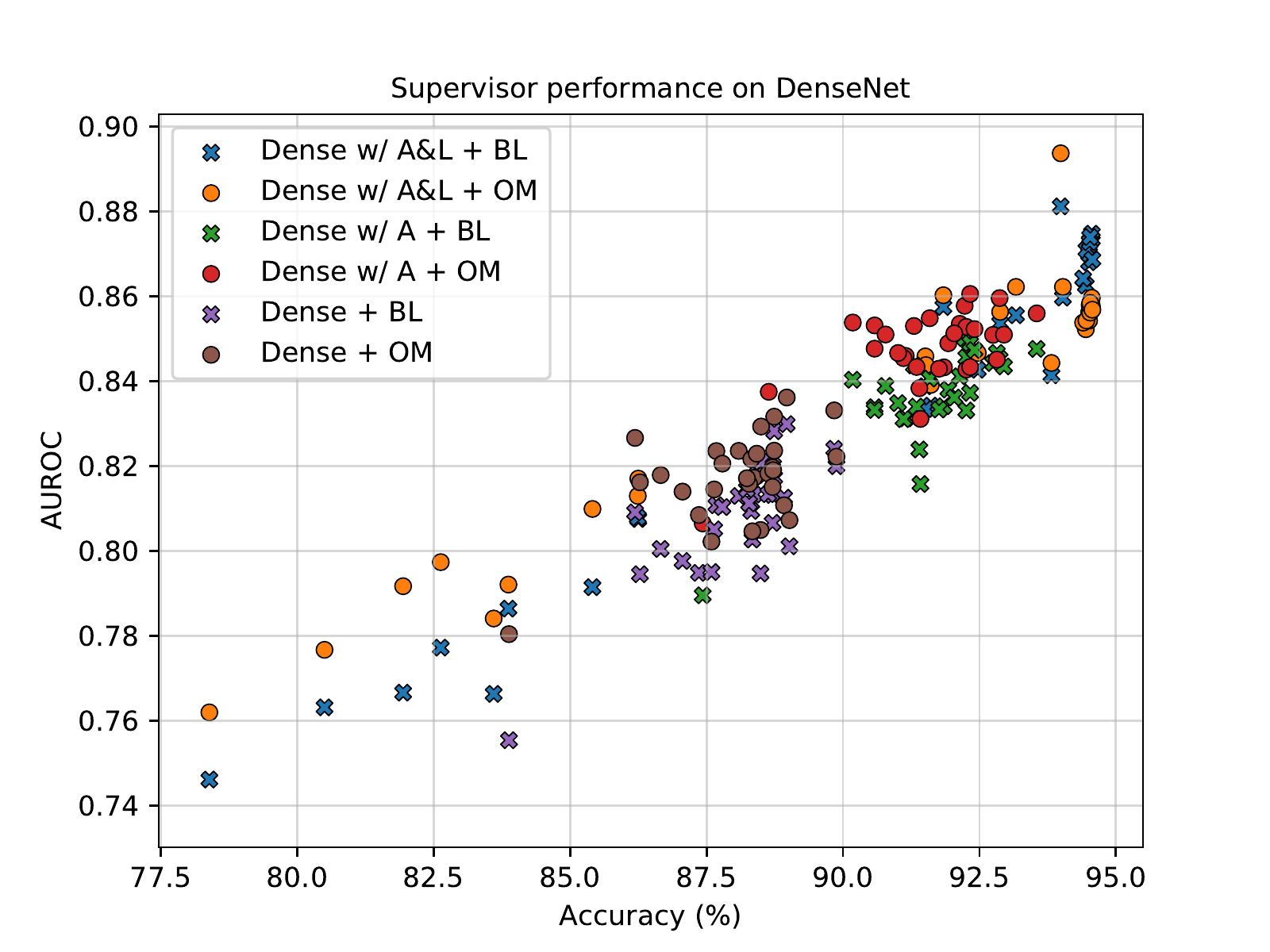}
    \caption{AUROC metric compared to the accuracy of the DenseNet network. With higher network accuracy, the supervisor can easier distinguish between in- and outliers.}
    \label{fig:dense-auroc-accuracy}
\end{figure}

To test the supervisors, we saved the model parameters every $10^{th}$ epoch during training for all training variants, as well as the best epoch based on the accuracy on the test set. For each epoch, the supervisors are then evaluated and analyzed. The metrics for each models' best performing epoch are presented in Table~\ref{tab:supervisor_best_results}. The table shows how each training improvement, i.e., approaches to support generalization, applied in this study improved the final accuracy. Coincidentally, we note that also the AUROC increased when the accuracy  increased. In fact, further investigation shows that the model accuracy and supervisor AUROC are connected in an almost linear relationship as can be seen in Fig.~\ref{fig:vgg-auroc-accuracy} and~\ref{fig:dense-auroc-accuracy}. This indicates that tools to improve the generalizability of the models also improve the ability to distinguish between in- and out-of-distribution samples, thus increasing the robustness of the system. 

When studying Figs.~\ref{fig:vgg-auroc-accuracy} and \ref{fig:dense-auroc-accuracy} further, it can be noted that certain locations in the plot are more dense that others. It is reasonable to expect epochs close-by to have similar results, thus have similar performance in the graphs. Since the supervisors are retrained for each epoch, some level of variance is achieved (e.g., around 90\% accuracy in Figs.~\ref{fig:vgg-auroc-accuracy} and~\ref{fig:dense-auroc-accuracy}).  This shows how subtle changes in the model can rapidly change the behavior of the supervisor to perform better/worse. 

Temporary bad performance of the supervisor can potentially be solved with better parameter settings of the supervisors (OpenMax in this case). In this study, a parameter search was conducted through a grid search for one of the epochs. The grid was set up so that the centre of the grid corresponded to the parameter settings in the original paper~\cite{bendale2016towards}. In total 1,000 different runs were conducted, and the best settings were exported and used for all epochs. Since the model changes after each epoch, this can potentially introduce another level of uncertainty for the supervisor, since the supervisors are not optimized at each epoch.

\begin{figure}[!t]
    \centering
    \includegraphics[width=1\linewidth]{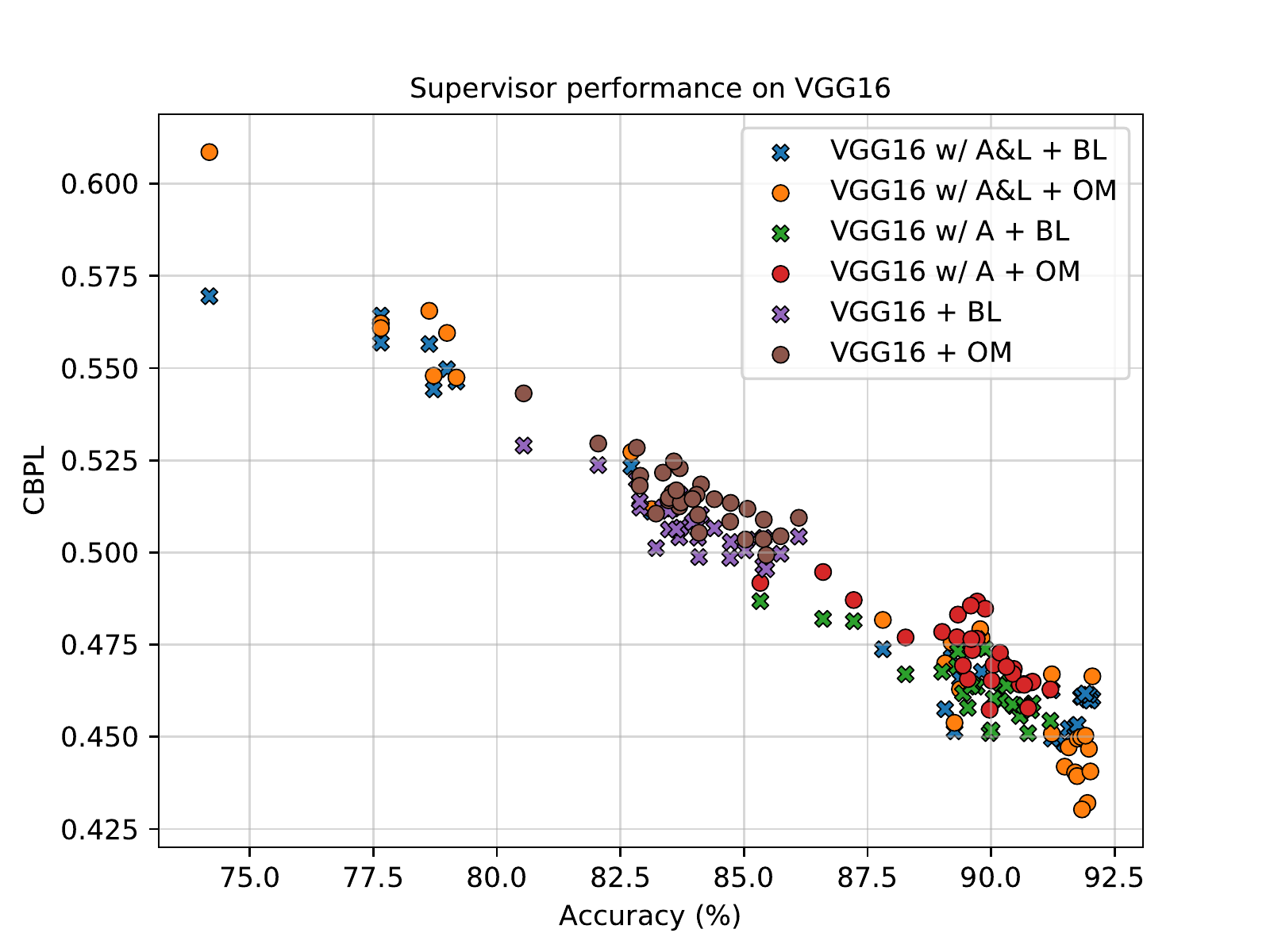}
    \caption{Coverage breakpoint at performance (CBPL) level compared to the test set accuracy of VGG16. Epochs with higher accuracy requires a stricted supervisor.}
    \label{fig:vgg_coverage}
\end{figure}

\begin{figure}[!t]
    \centering
    \includegraphics[width=1\linewidth]{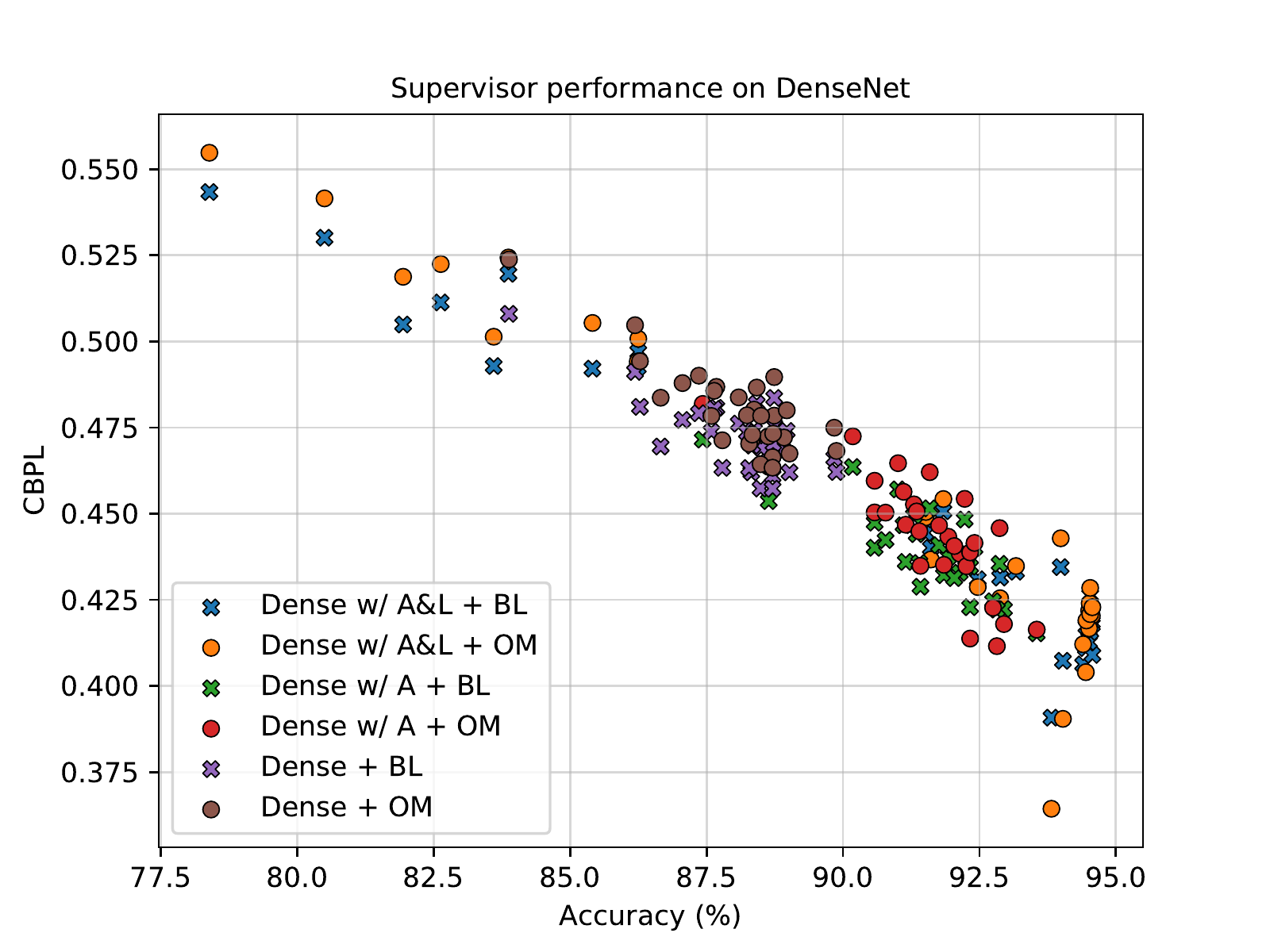}
    \caption{Coverage breakpoint at performance level (CBPL) compared to the test set accuracy of DenseNet. Epochs with higher accuracy requires a stricted supervisor.}
    \label{fig:densenet_coverage}
\end{figure}

When analyzing how the coverage breakpoint to achieve the same accuracy as during the training session compared to the actual accuracy of the DNN, the results also show linear behavior with some dense clusters, see Fig.~ \ref{fig:vgg_coverage} and \ref{fig:densenet_coverage}. Initially, it can be seen that an overal linear trend appears between the accuracy and coverage: As long as the accuracy of the DNN is increasing, the coverage to achieve same accuracy is decreased. Intuitively, this behavior is expected, since the margin error margin is smaller with higher accuracy, the supervisor requires to be stricter to ensure more potential false activations are rejected. This behavior is spotted for both of the supervisors. 

An important observation is that the ability of the OpenMax supervisor to detect out-of-distribution samples is superior compared to the Baseline algorithm for the majority of epochs. The OpenMax algorithm has only been tuned for one of the epochs, it still manages to perform well on the majority of epochs, even though it is unknown how much the actual parameters have changed between the different runs. The coverage breakpoint can however vary up towards 5\% for epochs with similar accuracy. This indicates that tiny parameter changes causes the stability of the network to vary. A good indication of a robust network+supervisor combination would indicate that these small changes would not affect the coverage in a such a large manner.   



\begin{figure}[!t]
    \centering
    \includegraphics[width=1\linewidth]{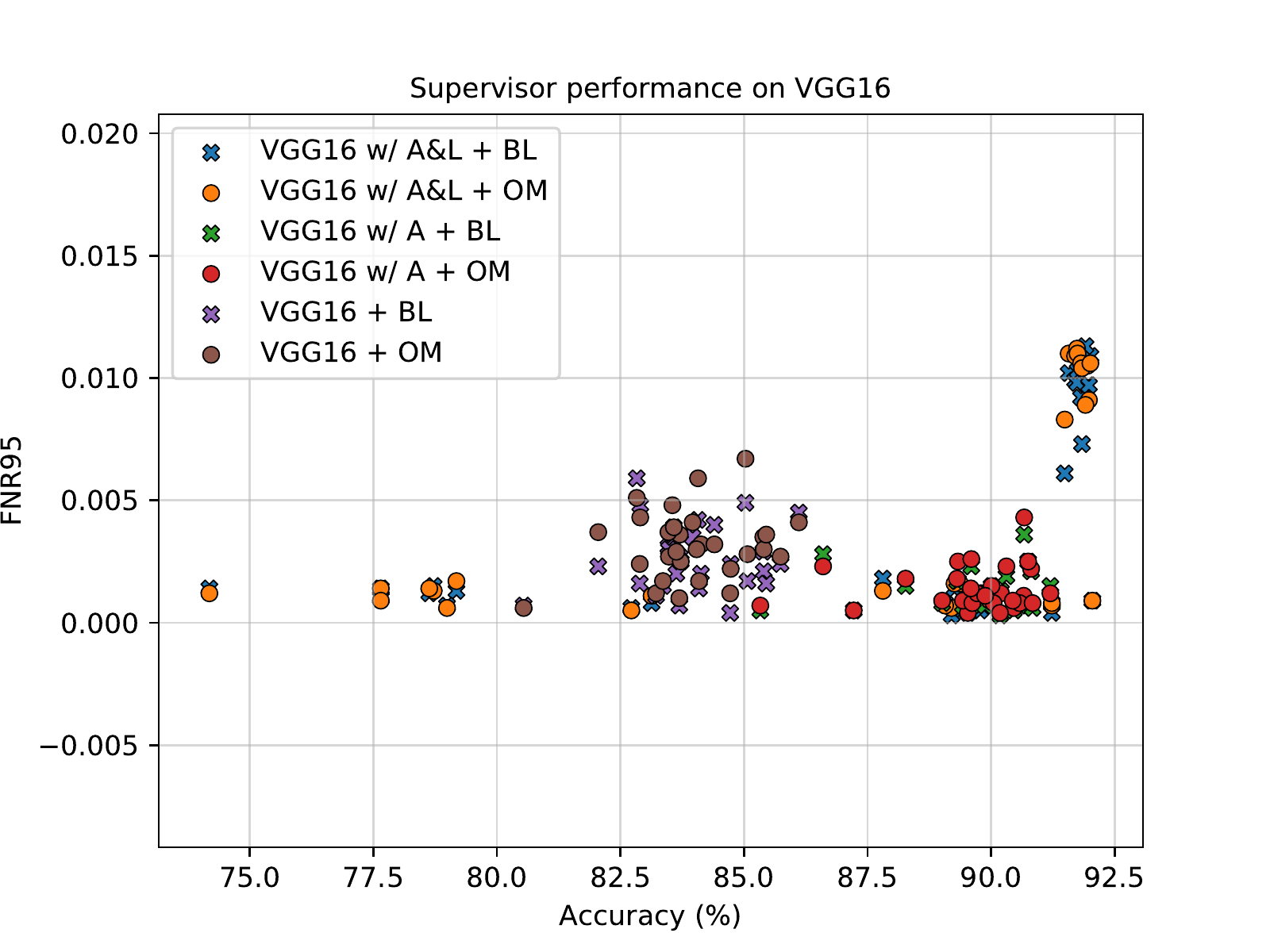}
    \caption{The false negative rate for all models compared the corresponding network test set accuracy on VGG16. Most missed samples are achieved by the most accurate networks, which suggest fine-tuning of the network increases the miss rate.}
    \label{fig:vgg_fnr95}
\end{figure}

\begin{figure}[!t]
    \centering
    \includegraphics[width=1\linewidth]{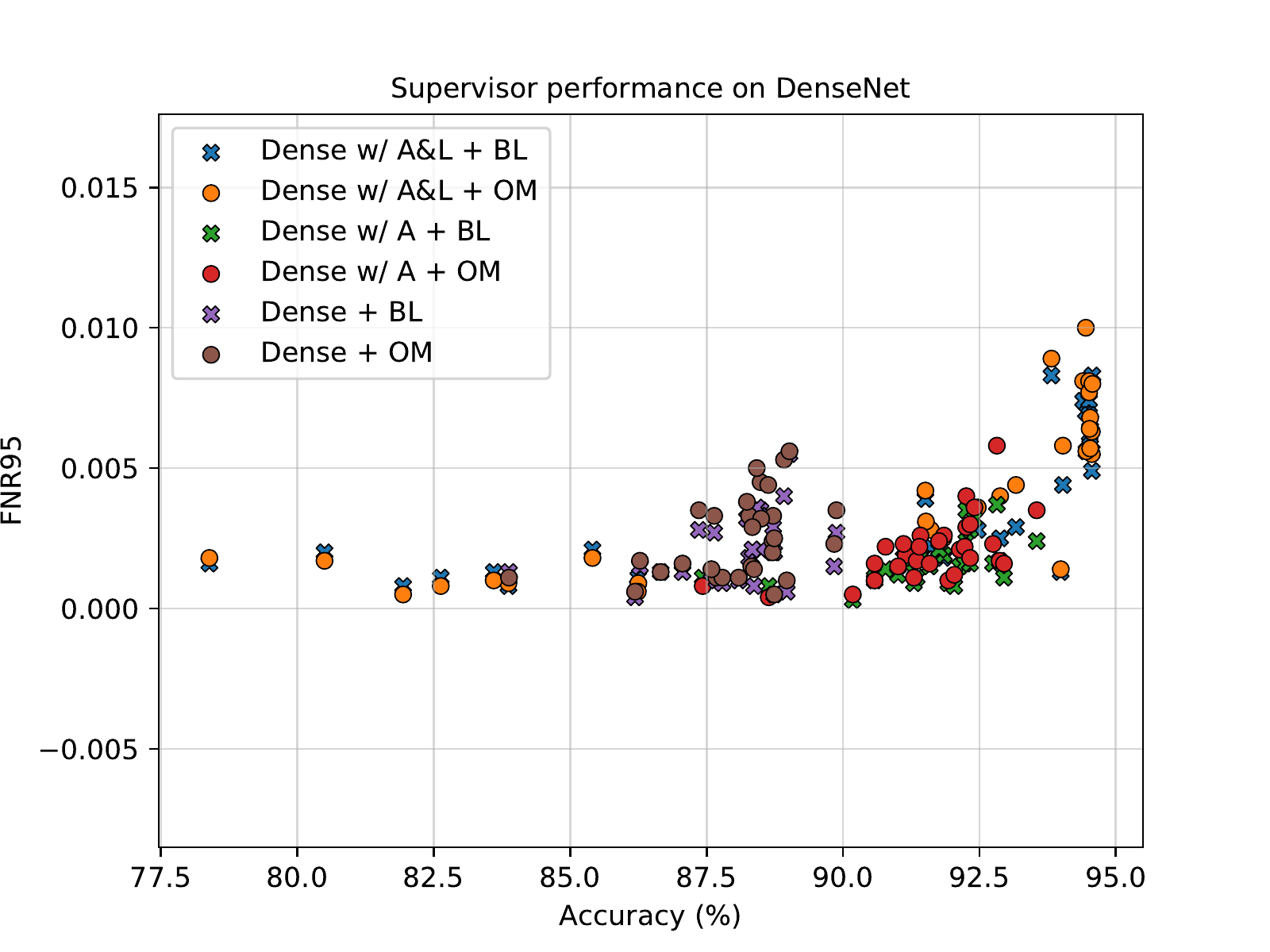}
    \caption{The false negative rate for all models compared the corresponding network test set accuracy on DenseNet. Most missed samples are achieved by the most accurate networks, which suggest finetuning of the network increases the miss rate.}
    \label{fig:densenet_fnr95}
\end{figure}

To further understand the robustness issue, we study how the false negative rate changes as the network gets more accurate. The false negative rate refers to how many samples are missed by the supervisor and hence, lower or close to zero is preferred. The results over all training epochs can be seen in Fig.~ \ref{fig:vgg_fnr95} and \ref{fig:densenet_fnr95}. Interestingly, the miss rate of outliers increases for the most accurate epochs. Intuitively, this is caused by two factors: The overfitting behavior of the fine-tuned networks causes the DNNs to be overconfident on samples it has not previously seen, as well as the fact that the datasets are not disjoint, thus certain samples look similar to the training domain, even though coming from the outlier distribution. This is not the case for earlier epochs, because the DNN has not yet overfit for these specific samples. 

To summarize, looking at Figs.~\ref{fig:vgg-auroc-accuracy}-\ref{fig:densenet_fnr95}, it is clear that improvements in generalization of the DNNs will increase the performance of the supervisors. Even with bad parameter initialization of the supervisors, the supervisor can exclude the majority of the out-of-distribution samples. Problems occur when the models are starting to fine-tune towards the training set, especially when allowed to use a smaller learning rate. When this happens, it is not guaranteed that the DNN generalizes any further, even though the accuracy might indicate improved results. This is further seen in the false negative rate plots, where the error increases when this fine-tuning starts, it can also be seen in the coverage reduction for higher accuracy epochs. The stability issue of the supervisors are shown for later epochs as well, where the results are dwindling downwards rather than improving, which raises the question of how to properly design and tune supervisors.


\section{Discussion and future work} \label{sec:disc}
Since more and more promising results gain attention and intended functionality becomes an increasingly discussed topic, the fact that DNNs can act overconfidently on input samples far from the training domain raises an issue. Even though several novel outlier detection methods have been published lately, as well as additional adversarial fooling methods, the lack of comparable results remains an issue. 

To advance the development of DNN supervisors, an ongoing issue is the comparability of these constructs, where in most cases the developers are training and testing on specific scenarios tailored for the task at hand, rendering it impossible to benchmark and compare in a proper way towards related work. In addition, proper ways to compare and benchmark would improve the way how parameterization of these supervisors would affect the behavior and instability. This is useful to see how they work in a safe environment compared to when exposed to adversarial or extreme case data. 

For safety critical applications, the need to distinguish between stable inputs (\textit{known safe states}), and uncertain states \textit{unknown states} is a major challenge that needs to be handled properly before deploying learning based systems into the real world, where all possible states are infeasible to record. Indeed, the size of the input space is one of the major challenges when engineering learning based systems~\cite{borg_safely_2019}, since it neither can be specified at design-time nor tested exhaustively before deployment.

While DNN supervisors have been proposed for perception in the literature before, this paper is the first to frame this safety mechanism within SOTIF. Based on the ``Identification and Evaluation of Triggering Events'' step in the ``Evaluate by Analysis'' of the SOTIF process (cf.~Fig.~9 in \cite{international_organization_for_standardization_iso_2019}), we argue that a functional restriction of computer vision is needed to mitigate the SOTIF related risks. Severe  hazards remain even in the presence of ISO~26262, since functional insufficiency might result in potentially fatal object detection. We propose to complement DNNs with supervisors to restrict the function, i.e., to introduce an additional means to perform out-of-distribution detection. We posit that supervisors could be a fundamental component in safety cage architectures tailored for computer vision in autonomous drive, and that a corresponding safety requirement should be traced to the DNN supervisor~\cite{borg_traceability_2017}.

However, DNN supervisors for perception in autonomous driving are not yet sufficiently understood. Before the SOTIF process can proceed with verification and validation of the intended function (including the proposed DNN supervisors), more research is needed. Open questions include how should the DNNs and their supervisors be trained and how does overfitting affect the results? In this paper, we show that improved generalization of the DNNs initially increases the performance of the associated supervisor. On the other hand, when the DNNs start overfitting, the performance of the supervisor deteriorates -- as made evident by diminishing false negative rates and coverage breakpoints. Further studies are needed to increase the external validity of our conclusion, involving additional DNN architectures and supervisors as well as different datasets.

\section{Conclusions} \label{sec:conc}
In this work we have studied the interplay between the performance of a Deep Neural Network (DNN) in terms of classification accuracy, and the ability of an anomaly detection system to identify out-of-distribution samples when supervising the DNN. For several DNNs, supervisors and performance metrics, we show that there is a clear correlation between network accuracy and the ability of the supervisor to detect out-of-distribution samples. Furthermore, we show that different aspects of the supervisor performance are affected differently as the network accuracy changes. There is for instance a notable trade-off between achieving a high AUROC and a low FNR95 score. It is important to realize that it is the application that determines the relative importance of the different metrics used to evaluate the supervisor. For a safety critical application, a low value of FNR95 (accepting a low number of anomalies) would probably be more important than a high AUROC value. 

The dependence of supervisor performance on the detailed structure of the DNN, and the choice of evaluation metrics, highlight the fact that the black box nature and high complexity of DNNs make them hard to analyze from a safety perspective. It is difficult to make general statements about a certain supervisor method without proper analysis of the details in both model training and implementation. We therefore encourage the scientific community to put more emphasis on the analysis of supervisor performance when new methods are presented.  

\section*{Acknowledgments}
This work was carried out within the SMILE II project financed by Vinnova, FFI, Fordonsstrategisk forskning och innovation under the grant number: 2017-03066, and partially supported by the Wallenberg AI, Autonomous Systems and Software Program (WASP) funded by Knut and Alice Wallenberg Foundation.

\bibliographystyle{IEEEtran}
\bibliography{smile}

\end{document}